\documentclass[11pt]{article}

\usepackage{acl2012}

\usepackage{times}
\usepackage{latexsym}
\usepackage{amsmath}
\usepackage{multirow}
\usepackage{url}
\usepackage{fancyvrb}
\usepackage{paralist}

\usepackage{multicol}

\title{An Inventory of Preposition Relations}

\author{Vivek Srikumar \and Dan Roth \\
  University of Illinois, Urbana-Champaign\\
  Urbana, IL. 61801.\\
  \{vsrikum2, danr\}@illinois.edu
}
\date{}
\newcommand{\relation}[1]{{\texttt {#1}}}
\newcommand{\example}[1]{{\em #1}}

\begin{document}
\maketitle

\begin{abstract}
  We describe an inventory of semantic relations that are expressed by
  prepositions. We define these relations by building on the word
  sense disambiguation task for prepositions and propose a mapping
  from preposition senses to the relation labels by collapsing
  semantically related senses across prepositions.
\end{abstract}

\section{Introduction}
\label{sec:intro}
This document defines a taxonomy of relations expressed by
prepositions and serves as a companion paper for \cite{SrikumarRo13},
which presents a computational model for predicting them. These
relations are defined by collapsing related preposition senses from
the inventory defined by the Preposition Project
\cite{LitkowskiHa05}\footnote{The preposition sense inventory can be
  accessed online at \url{http://www.clres.com/prepositions.html}.}.
The sense inventory, which is based on the definitions of prepositions
in the Oxford Dictionary of English, treats each dictionary definition
of the preposition as a separate sense and has been used by the 2007
shared task of preposition sense disambiguation \cite{LitkowskiHa07}.

By grouping semantically related senses across prepositions into
relations, we define 32 relation labels covering the 34
prepositions\footnote{We cover the transitive usages of the following
  prepositions: about, above, across, after, against, along, among,
  around, as, at, before, behind, beneath, beside, between, by, down,
  during, for, from, in, inside, into, like, of, off, on, onto, over,
  round, through, to, towards, with.} considered by the Preposition
Project. 
In addition to defining each preposition sense, the Preposition
Project also indicates related prepositions. The sense definitions and
the related prepositions let us identify senses that carry the same
meaning across prepositions to define the relation labels. We followed
this with a manual cleanup phase. Some senses do not cleanly align
with a single relation because the definitions include idiomatic or
figurative usage. For example, the sense {\bf in:7(5)} of the
preposition \example{in} includes both spatial and figurative notions
of the spatial sense (that is, both \example{in London} and
\example{in a film}). In such cases, we sampled 20 examples from the
SemEval 2007 training set and assigned the relation label based on
majority.
One of the labels is a catch-all category, \relation{Other}, which
consists of many infrequent senses of prepositions and is not expected
to be a semantically coherent class.

Two native speakers of English annotated 200 sentences from the
SemEval training corpus using only the definitions of the labels (as
listed in this document) as the annotation guidelines. This helps us
validate the mapping from senses to relations. the We measured Cohen's
kappa coefficient \cite{Cohen1960} between the annotators to be 0.75
and also between each annotator and the original corpus to be 0.76 and
0.74 respectively.

Section \ref{sec:preposition-relations-list} defines all the relations
and provides examples of prepositional phrases that exhibit
them. Section \ref{sec:statistics} lists the number of training and
test examples obtained by converting the SemEval 2007 preposition
sense data using the mapping defined in Section
\ref{sec:preposition-relations-list}.

\section{Preposition relations}
\label{sec:preposition-relations-list}

For each relation that is defined in this section, we list the set of
prepositions and their senses that are included as part of this
relation. The sense identifiers refer to those used by the the SemEval
2007 shared task.
Senses that are infrequent according to the training data for the
shared task (that is, seen fewer than five times) are marked with an
asterisk.

\subsection{Activity}
\paragraph{Definition:}  Describes the relationship between some entity and an activity, an ordeal or a process that can be a verbal noun or a gerund. Note that the object of the preposition is the activity here.
\paragraph{Examples:} \hspace*{\fill}
\begin{itemize}
\item he is into jet-skiing
\item good at boxing
\item prudent in planning
\item out on errands
\item prelude to disaster
\item struggle through it
\end{itemize} 

{\small


\begin{tabular}{|l|l|}
\hline
{\bf Prep.} & {\bf Senses} \\ \hline
at&8(4a) \\
in&11(8) \\
into&9(9)$^*$ \\
on&19(9)$^*$ \\
through&8(2a)$^*$ \\
to&10(4a) \\
\hline\end{tabular}}
\paragraph{Comments:}   If there is a change of state and the object of the preposition is the final state, the preposition indicates the relation \relation{EndState}.
\subsection{Agent}
\paragraph{Definition:}  The object of the preposition is the agent of the action indicated by the attachment point. This primarily covers the use of the preposition {\em by} in a passive construction.
\paragraph{Examples:} \hspace*{\fill}
\begin{itemize}
\item understood by the customers
\item the address by the officer
\item recent discovery by nutritionists
\item allegations from the FDA
\end{itemize} 

{\small


\begin{tabular}{|l|l|}
\hline
{\bf Prep.} & {\bf Senses} \\ \hline
by&1(1)$^*$ 2(1a) 3(1b) \\
from&12(9)-1 \\
\hline\end{tabular}}
\paragraph{Comments:}   This relation does not include prepositional phrases like {\em sonnet by Shakespeare} because {\em Shakespeare} is the agent of a verb such as write or compose, and not the word {\em sonnet}. This type of creator-creation relation will be labeled as \relation{Source}.
\subsection{Attribute}
\paragraph{Definition:}  The object of the preposition indicates some attribute of either the governor or the subject/object of the governor in case of verb-attached prepositions.
\paragraph{Examples:} \hspace*{\fill}
\begin{itemize}
\item interpreted their action as betrayal
\item sell at a loss
\item breakdown by age
\item what is meant by ``fair''?
\item call by last name
\item paint made from resin
\item blinked in astonishment
\item dressed in a sweater
\item Mozart's Piano Concerto in E flat
\item house built of bricks
\item smile of delight
\item man with glasses
\item blouse with white collar
\end{itemize} 

{\small


\begin{tabular}{|l|l|}
\hline
{\bf Prep.} & {\bf Senses} \\ \hline
as&1(1) \\
at&7(4)$^*$ \\
by&15(3c) 6(2a)$^*$ 7(2b)$^*$ \\
from&9(6) \\
in&1(1)-1 10(7a)$^*$ 6(4a) 9(7)-1 \\
of&17(8) 6(3)-1 \\
with&2(2) 3(2a) \\
\hline\end{tabular}}
\paragraph{Comments:}   \begin{enumerate} \item Prepositions where the governor indicates an attribute are classified as \relation{Possessor}. \item If the object of the preposition does not indicate an attribute, this relation does not apply. \end{enumerate}
\subsection{Beneficiary}
\paragraph{Definition:}  The object of the preposition is a beneficiary of the action indicated by the preposition's governor.
\paragraph{Examples:} \hspace*{\fill}
\begin{itemize}
\item fought for Napoleon
\item vote for independence
\item present for you
\item be charming to them
\item clean up after him
\item throw weight behind candidate
\end{itemize} 

{\small


\begin{tabular}{|l|l|}
\hline
{\bf Prep.} & {\bf Senses} \\ \hline
after&4(1c)$^*$ \\
behind&5(3)$^*$ \\
for&1(1) 3(3) \\
to&8(3)-1 \\
\hline\end{tabular}}
\paragraph{Comments:}   This relation includes examples like {\em the power behind the throne}, where the phrase indicates that the object of the preposition gains some advantage.
\subsection{Cause}
\paragraph{Definition:}  The object indicates a cause for the governor.
\paragraph{Examples:} \hspace*{\fill}
\begin{itemize}
\item in bed with flu
\item shaking with anger
\item wisdom comes with age
\item died of cancer
\item suffering from asthma
\item admire for bravery
\item tired after work
\item disapproval at the behavior
\item agony behind the decision
\end{itemize} 

{\small


\begin{tabular}{|l|l|}
\hline
{\bf Prep.} & {\bf Senses} \\ \hline
after&1(1)-1 \\
at&11(6)-1 \\
for&6(5) \\
from&12(9) \\
of&16(7b) \\
with&10(7a)$^*$ 11(7b) 12(7c) \\
\hline\end{tabular}}
\subsection{Co-Participants}
\paragraph{Definition:}  The preposition indicates a choice, sharing or differentiation and involves multiple participants, represented by the object.
\paragraph{Examples:} \hspace*{\fill}
\begin{itemize}
\item drop in tooth decay among children
\item divide his kingdom among them
\item links between science and industry
\item the difference between income and expenditure
\item choose between two options
\end{itemize} 

{\small


\begin{tabular}{|l|l|}
\hline
{\bf Prep.} & {\bf Senses} \\ \hline
among&3(3) 4(4)$^*$ \\
between&4(4) 6(4b) 8(5)$^*$ 9(5a)$^*$ \\
\hline\end{tabular}}
\paragraph{Comments:}   This is different from \relation{Participant/Accompanier} where the governor and the object indicate the two entities that are associated. Here, the object specifies all the participants.
\subsection{Destination}
\paragraph{Definition:}  The object of the preposition is the destination of the motion indicated by the governor.
\paragraph{Examples:} \hspace*{\fill}
\begin{itemize}
\item leaving for London tomorrow
\item put coal in the bath
\item tucked the books inside his coat
\item Sara got into her car
\item crashed into a parked car
\item the road led into the village
\item walking to the shops
\end{itemize} 

{\small


\begin{tabular}{|l|l|}
\hline
{\bf Prep.} & {\bf Senses} \\ \hline
for&7(6)$^*$ \\
in&2(1a) \\
inside&2(1a) \\
into&1(1) 2(2) 3(3) \\
to&1(1) \\
\hline\end{tabular}}
\paragraph{Comments:}   \begin {enumerate} \item One boundary case which is not a \relation{Destination} is that of a motion that indicates that two things are attached or lined. Such cases indicate the relation \relation{Participant/Accompanier}. For example, {\em fasten the insulator to the frame} might indicate motion of the insulator to the frame, but Participant/Accompanier is a better choice because it indicates that the two objects are linked to each other. \item If there is not physical motion, but merely a change of state, the relation \relation{EndState} is preferred. \end{enumerate}
\subsection{Direction}
\paragraph{Definition:}  The prepositional phrase (that is, the preposition along with the object) indicates a direction that modifies the action which is expressed by the governor.
\paragraph{Examples:} \hspace*{\fill}
\begin{itemize}
\item crept up behind him
\item shut the door after her
\item driving along the road
\item drive by the house
\item tears streaming down her face
\item wander down the road
\item roll off the bed
\item drive towards the house
\item swim with the current
\end{itemize} 

{\small


\begin{tabular}{|l|l|}
\hline
{\bf Prep.} & {\bf Senses} \\ \hline
after&5(2) 6(2a)$^*$ \\
along&1(1) \\
behind&4(2a) \\
by&19(5a)$^*$ \\
down&1(1) 2(1a) 3(1b) \\
off&1(1) \\
towards&1(1) \\
with&15(9)$^*$ \\
\hline\end{tabular}}
\paragraph{Comments:}   \begin{enumerate} \item This relation sometimes shares meaning with the relation \relation{Location}. \item This relation does not include metaphorical meaning and almost always has a spatial sense \end{enumerate}
\subsection{EndState}
\paragraph{Definition:}  The object of the preposition indicates the resultant state of an entity that is undergoing change. The governor of the preposition is usually a verb or noun denoting transformation.
\paragraph{Examples:} \hspace*{\fill}
\begin{itemize}
\item protest turned into a violent confrontation
\item forced the club into a special meeting
\item profits dropped to \$75 million
\item she was moved to tears
\item smashed to smithereens
\item return towards traditional Keynesianism
\end{itemize} 

{\small


\begin{tabular}{|l|l|}
\hline
{\bf Prep.} & {\bf Senses} \\ \hline
into&6(6) 7(7) \\
to&3(1b)$^*$ 5(2)$^*$ 6(2a) \\
towards&2(1a)$^*$ \\
\hline\end{tabular}}
\paragraph{Comments:}   \begin{enumerate} \item This relation requires some change of state, where the object of the preposition indicates the final state at the end of the transformation. \item This relation includes cases like {\em level fell to 3 feet}, where the object of the preposition is a numeric quantity. See note for \relation{Numeric}. \end{enumerate}
\subsection{Experiencer}
\paragraph{Definition:}  The object of the preposition denotes the entity that is target or subject of the action, thought or feeling that is usually indicated by the governor.
\paragraph{Examples:} \hspace*{\fill}
\begin{itemize}
\item focus attention on her
\item he blamed it on her
\item he was warm toward her
\item felt angry towards him
\end{itemize} 

{\small


\begin{tabular}{|l|l|}
\hline
{\bf Prep.} & {\bf Senses} \\ \hline
on&11(5) 11(5)-1 \\
towards&4(2) 4(2)-1 \\
\hline\end{tabular}}
\paragraph{Comments:}   There are some examples where the relation might overlap with \relation{Topic}. The sense on:11(5) includes both phrases like {\em focus attention on the her}, which conform to the definition of \relation{Experiencer}, and also {\em deliberate on the matter}, which are better suited to be labeled \relation{Topic}. However, in this annotation, the sense is labeled as \relation{Experiencer}.
\subsection{Instrument}
\paragraph{Definition:}  The object of the preposition indicates the means or the instrument for performing an action that is typically the governor.
\paragraph{Examples:} \hspace*{\fill}
\begin{itemize}
\item hold at knifepoint
\item provide capital by borrowing
\item banged his head on the beam
\item voice over the loudspeaker
\item heard through the grapevine
\item cut the fish with a knife
\item fill the bowl with water
\end{itemize} 

{\small


\begin{tabular}{|l|l|}
\hline
{\bf Prep.} & {\bf Senses} \\ \hline
at&11(6)$^*$ \\
by&5(2) \\
on&3(1b) \\
over&15(6)$^*$ 15(6)-1$^*$ \\
through&13(5a) \\
with&4(3) 5(3a) \\
\hline\end{tabular}}
\paragraph{Comments:}   \begin{enumerate} \item This relation includes both the means of performing actions and instruments. Though there are subtle differences between them, they are collapsed into one category to avoid sparse labels. \item The sense at:11(6) refers to idiomatic uses, where the object of the preposition can be words {\em knifepoint} or {\em gunpoint}. These are marked as \relation{Instrument} because they indicate that the action was performed with a knife or a gun. \end{enumerate}
\subsection{Location}
\paragraph{Definition:}  The prepositional phrase indicates a locative meaning. This relation includes both physical and figurative aspects of location. That is, both {\em left it in the cupboard} and {\em left it in her will} are treated as \relation{Location}s. (See note below.)
\paragraph{Examples:} \hspace*{\fill}
\begin{itemize}
\item look about the room
\item the cable runs above the duct
\item the hills above the capital
\item bruises above both eyes
\item travel across Europe
\item parked along the grass
\item hidden among the roots
\item large depots around the country
\item live at Conway house
\item she stood before her
\item kept behind the screens
\item labyrinths beneath Moscow
\item sat beside her
\item border between the countries
\item discovered by the roadside
\item going down the pub
\item see them from here
\item living in London
\item waiting inside the house
\item wind blowing into your face
\item north of Watford
\item street off Whitehall
\item camp on the island
\item copy onto a disk
\item flames over the city
\item went round the house
\item stepped through the doorway
\item forty miles to the south of the site
\end{itemize} 

{\small


\begin{tabular}{|l|l|}
\hline
{\bf Prep.} & {\bf Senses} \\ \hline
about&3(2) 3(2)-1 4(3)$^*$ \\
above&1(1)$^*$ 2(1a) 3(1b)$^*$ 4(2) \\
across&1(1) 2(2) \\
along&3(2) \\
among&1(1) \\
around&1(1) 3(2) 4(3) 4(3)-1 5(4) \\
at&1(1) \\
before&2(2) 3(2a) \\
behind&1(1) 3(2) \\
beneath&1(1) 2(1a) 3(2) \\
beside&1(1) \\
between&1(1) \\
by&18(5) \\
down&4(1c)$^*$ \\
from&8(5) \\
in&1(1) 7(5) \\
inside&1(1) 3(1b) \\
into&4(4) \\
of&8(4)$^*$ \\
off&2(2)$^*$ 3(2a) \\
on&2(1a)$^*$ 7(2) \\
onto&1(1) \\
over&1(1)$^*$ 11(4) 12(4a)$^*$ 13(4b) 2(1a) 3(1b) \\
& 4(2) \\
round&1(1) 3(2) 4(2a) 5(3) 6(3a) 8(4) \\
through&1(1) 12(5)-1$^*$ 2(1a) 3(1b) 4(1c) 5(1d) \\
& 6(1e)$^*$ \\
to&2(1a) \\
\hline\end{tabular}}
\paragraph{Comments:}   Categorizing preposition senses as \relation{Location}s is not always easy. Some senses may indicate multiple relations based on the specific sentence and the sense could have ideally been split into two different labels. For example, at:1(1) includes both destinations (e.g. {\em they stopped at a small trattoria}) and locations (e.g. they live at Conway House). In such ambiguous cases, the sense has been categorized as \relation{Location}. The preposition {\em in}, in the sense in:7(5) includes figurative meanings like {\em read it in a book} and has been included here because this is the closest meaning. The same holds for inside:3(1b), which includes phrases like {\em anger simmered inside me}. An alternative relation could be \relation{PartWhole}. Some examples that are not a \relation{Location} relation are: \begin{enumerate} \item when there is a comparison to a norm or a number (eg. {\em the food was above average}), which is an \relation{Other} relation; \item when there is a notion of something being attached to or coming into physical contact with something else, where \relation{Participant/Accompanier} is a better fit; \item when one argument of the preposition is indicated to be a member of the other, where \relation{PartWhole} is a better fit; \item when one object is physically supported by a surface or another object, where \relation{PhysicalSupport} is a better choice. \end{enumerate}
\subsection{Manner}
\paragraph{Definition:}  The prepositional phrase indicates the manner in which an action is performed. The action is typically the governor of the preposition.
\paragraph{Examples:} \hspace*{\fill}
\begin{itemize}
\item frame the definition along those lines
\item blinked in astonishment
\item disappear in a flash
\item plummet like a dive-bomber
\item obtained through fraudulent means
\item shout with pleasure
\end{itemize} 

{\small


\begin{tabular}{|l|l|}
\hline
{\bf Prep.} & {\bf Senses} \\ \hline
along&4(3)$^*$ \\
in&5(4) 6(4a)-1 \\
like&1(1) 2(1a) 3(1b) 4(1c)$^*$ 5(1d) 6(2)$^*$ \\
through&12(5) \\
with&7(5) \\
\hline\end{tabular}}
\paragraph{Comments:}   There may be some overlap of meaning between \relation{Manner}, \relation{Attribute} and \relation{Cause}, stemming from the preposition sense in:5(4). For example, consider the following phrases that are included in this sense: {\em pursed her lips in a silent whistle}, {\em shrugged in embarrassment} and {\em seething in outrage}.
\subsection{MediumOfCommunication}
\paragraph{Definition:}  The prepositional phrase indicates the medium or language of some form of communication or idea. The object is, in a general sense, a `mode of communication'. This includes languages (eg. {\em say it in French}), media like TV or the Internet (eg. {\em saw it on a website}), or specific instances of these (eg. {\em saw it on the Sopranos}).
\paragraph{Examples:} \hspace*{\fill}
\begin{itemize}
\item say it in French
\item put your idea down on paper
\item saw the new series on TV
\end{itemize} 

{\small


\begin{tabular}{|l|l|}
\hline
{\bf Prep.} & {\bf Senses} \\ \hline
in&9(7) \\
on&12(6) 13(6a) \\
\hline\end{tabular}}
\paragraph{Comments:}   In boundary cases between \relation{MediumOfCommunication} and \relation{Attribute} (as in {\em Shakespeare's plays in comic book form}), the former is a better choice if it is a specific case of \relation{Attribute}. The same argument holds for the boundary cases with \relation{Manner}.
\subsection{Numeric}
\paragraph{Definition:}  The object of the preposition indicates a numeric quantity (age, price, percentage, etc).
\paragraph{Examples:} \hspace*{\fill}
\begin{itemize}
\item driving at 50mph
\item missed the shot by miles
\item crawled for 300 yards
\item a boy of 15
\end{itemize} 

{\small


\begin{tabular}{|l|l|}
\hline
{\bf Prep.} & {\bf Senses} \\ \hline
at&5(3) 6(3a)$^*$ \\
between&3(3) \\
by&12(3) 13(3a)$^*$ 16(3d)$^*$ \\
for&10(8a) 13(11) 14(12)$^*$ \\
into&8(8)$^*$ \\
of&4(2) 5(2a)$^*$ \\
on&23(13)$^*$ \\
to&11(4b)$^*$ 12(4c)$^*$ \\
\hline\end{tabular}}
\paragraph{Comments:}   \begin{enumerate} \item Several senses included in this relation do not have any examples in the training set. \item The prepositions in phrases such as {\em drop from \$105 million to \$75 million} are not labeled as \relation{Numeric}. Instead, the {\em from} is labeled as \relation{StartState} and the {\em to} is labeled as \relation{EndState}. \end{enumerate}
\subsection{ObjectOfVerb}
\paragraph{Definition:}  The object of the preposition is an object of the verb or the nominalization that is the governor of the preposition. This includes cases like {\em construction of the library}, where the object of the preposition is an object of the underlying verb.
\paragraph{Examples:} \hspace*{\fill}
\begin{itemize}
\item inquired after him
\item chase after something
\item sipped at his coffee
\item considerations for the future
\item saved from death
\item the wedding of his daughter
\item it was kind of you
\item she tells of her marriage
\item presided over the meeting
\item scan through document
\item a threat to world peace
\item a grant towards the cost
\item cross with her.
\end{itemize} 

{\small


\begin{tabular}{|l|l|}
\hline
{\bf Prep.} & {\bf Senses} \\ \hline
after&7(3) \\
at&10(5a) 9(5) \\
for&2(2) 2(2)-1 \\
from&11(8) \\
of&11(6) 12(6a) 13(6b) 14(7) 15(7a) \\
on&9(3a) \\
over&6(2b) \\
round&7(3b)$^*$ \\
through&10(3) 10(3)-1$^*$ \\
to&14(6) \\
towards&5(3)$^*$ \\
with&15(9)-1 4(3)-1 9(7), \\
\hline\end{tabular}}
\paragraph{Comments:}   \begin{enumerate} \item Many of the other relations can be considered to be an object of the verb in question. However, if the sense of a preposition was better suited to a different relation, that relation has been chosen. With this caveat, we expect the object of the preposition to be a core argument for the governor according to the PropBank or the NomBank schemes.  \item This relation includes the senses for:2(2), of:13(6b), to:14(6) and with:9(7) which are very vaguely defined in the Preposition Project.  This contributes to some noise in the data. \item It is possible that some of the senses from the \relation{Other} category could be grouped into this relation, if there is more evidence in the data. \end{enumerate} Some examples which are not \relation{ObjectOfVerb} are: \begin{inparaenum} \item when there is a notion of connection or physical contact between entities, \relation{Participant/Accompanier} is a better fit, and \item when the object of the preposition is the target of an action or is affected by it, \relation{Recipient} or \relation{Beneficiary} is preferred over this label. \end{inparaenum}
\subsection{Opponent/Contrast}
\paragraph{Definition:}  This relation indicates a collision, conflict or contrast and the object of the preposition refers to one or more entities involved.
\paragraph{Examples:} \hspace*{\fill}
\begin{itemize}
\item fight against crime
\item gave evidence against him
\item the match against Somerset
\item turned up his collar against the wind
\item the wars between Russia and Poland
\item fees are distinct from expenses
\item fought with another man
\end{itemize} 

{\small


\begin{tabular}{|l|l|}
\hline
{\bf Prep.} & {\bf Senses} \\ \hline
against&1(1) 2(1a) 3(1b) 6(2b) \\
between&5(4a) 7(4c) \\
from&14(11) \\
with&6(4) \\
\hline\end{tabular}}
\paragraph{Comments:}  The main distinguishing criterion is that there are one or more antagonists in a conflict or in opposition. This includes cases like {\em pulled up the collar against the wind}. Some near misses include: \begin{enumerate} \item when there is no explicit conflict, the relation \relation{Co-Participants} is a better fit, and \item when there is a hint of a conflict, without mention of the entities involved (like {\em turn away from appeasement} and {\em stagger from crisis to crisis}), the relation \relation{Source} would be a better fit. \end{enumerate}
\subsection{Other}
\paragraph{Definition:}  This is the catch-all category, which includes infrequent senses of prepositions that do not fit any other category. A preposition being labeled as \relation{Other} does not mean that there is no relation expressed by it. Neither does it indicate that there is no shared meaning across prepositions that can be abstracted. In most cases, there is such little support from the training set for this label that either adding it to an existing relation or creating a new one cannot be justified.
\paragraph{Examples:} \hspace*{\fill}
\begin{itemize}
\item at a level above the people
\item married above her
\item heard above the din
\item a drawing after Millet's The Reapers
\item health comes after housing
\item named her Pauline after her mother
\item gritted his teeth against the pain
\item odds were 5-1 against them
\item money loaned against the property
\item benefits weighed against the costs
\item picked up tips along the way
\item been ill as a child
\item placed duty before everything
\item years behind them
\item he was rather beneath the princess
\item tragic life beneath the gloss
\item felt clumsy beside her
\item right by me
\item swear by God
\item `F' is for fascinating
\item swap for that
\item tall for her age
\item works like Animal Farm
\item he is on morphine
\item drinks are on me
\item married for over a year
\item the director is over him
\item he smiled to her astonishment
\item leave it with me
\end{itemize} 

{\small


\begin{tabular}{|l|l|}
\hline
{\bf Prep.} & {\bf Senses} \\ \hline
above&5(2a)$^*$ 6(2b)$^*$ 7(2c)$^*$ 8(2d) 9(3) \\
after&10(5a)$^*$ 8(4)$^*$ 9(5)$^*$ \\
against&4(2) 5(2a)$^*$ 7(2c)$^*$ 8(3)$^*$ 9(3a)$^*$ \\
along&2(1a)$^*$ \\
as&2(2)$^*$ \\
before&4(3)$^*$ \\
behind&2(1a)$^*$ 6(3a)$^*$ 8(5)$^*$ 9(6)$^*$ \\
beneath&4(2a)$^*$ 5(2b)$^*$ 6(2c)$^*$ \\
beside&2(1a)$^*$ 3(2)$^*$ \\
by&21(7)$^*$ 22(8)$^*$ \\
for&11(9) 8(7) 9(8) \\
inside&4(1c)$^*$ \\
like&7(3)$^*$ \\
on&20(10)$^*$ 21(11)$^*$ 22(12)$^*$ 9(3a)-1$^*$ \\
over&10(3)$^*$ 5(2a)$^*$ 7(2c)$^*$ 8(2d)$^*$ 9(2e)$^*$ \\
to&15(7)$^*$ 16(8)$^*$ 7(2b)$^*$ \\
with&8(6)$^*$ \\
\hline\end{tabular}}
\paragraph{Comments:}   Even though this label includes many senses, as we see in \ref{tab:num-train-test}, this relation is has a very small representation in the data. Being a very infrequent category, this label would apply only when none of the other relations hold.
\subsection{PartWhole}
\paragraph{Definition:}  This relation indicates that one argument is a part or member of another. It includes two distinct cases: (1) the governor of the preposition (or the subject of the governor, if the governor is a verb) is a part of the object, {\em and} (2) the governor is a number, a partitive noun or a container and the object is a group or substance that is modified by the governor.
\paragraph{Examples:} \hspace*{\fill}
\begin{itemize}
\item see a friend among them
\item sleeve of the coat
\item a slice of the cake
\item group of monks
\item cup of soup
\end{itemize} 

{\small


\begin{tabular}{|l|l|}
\hline
{\bf Prep.} & {\bf Senses} \\ \hline
among&2(2) \\
in&12(9)$^*$ \\
of&1(1)$^*$ 2(1a) 3(1b) 3(1b)-1 \\
\hline\end{tabular}}
\paragraph{Comments:}   \begin{enumerate} \item The two cases were clubbed together because both indicate a notion of an object or substance being divided into parts or groups. This is distinct from the \relation{Possessor} relation, where neither argument is divided to define the other. \item This label indicates all quantifiers and partitive nouns that are connected to entities via the preposition. For example, {\em hundreds of people}, {\em a piece of cake}, {\em many of the protesters}. \item This label also includes cases where the object of the preposition is a set of entities (or a container) and the governor is an element of that set (or the contents of the container), as in the example, {snakes are among the most feared animals}. \end{enumerate}
\subsection{Participant/Accompanier}
\paragraph{Definition:}  The object of the preposition indicates an entity which accompanies another entity or participates in a relation with another entity, which is typically indicated by either the governor of the preposition or the subject of the governor.
\paragraph{Examples:} \hspace*{\fill}
\begin{itemize}
\item stick the drawings onto a large map
\item he is married to Emma
\item a map pinned to the wall
\item his marriage with Emma
\end{itemize} 

{\small


\begin{tabular}{|l|l|}
\hline
{\bf Prep.} & {\bf Senses} \\ \hline
by&11(2f)$^*$ \\
onto&3(3) \\
to&13(5) 9(4) \\
with&1(1) \\
\hline\end{tabular}}
\paragraph{Comments:}   \begin{enumerate} \item In this relation, the object indicates only one of the participating entities. In contrast, in the \relation{Co-Participants} relation, the object defines all the participating entities, typically in the form of a plural or a conjunction. \item The sense by:11(2f) is included here based on the notes on the treatment of the preposition in the Preposition Project. However, there are no examples to support this inclusion, and this may be moved to the relation \relation{Other}. \item All cases where something is joined to something else should be treated as \relation{Participant/Accompanier}. There might be some confusion between this label and \relation{ObjectOfVerb}, \relation{Destination} or \relation{Location}. Whenever there is the notion of connection between entities, this label is a better choice.  \end{enumerate}
\subsection{PhysicalSupport}
\paragraph{Definition:}  The object of the preposition is physically in contact with and supports the governor (if it is an entity) or the subject of the governor (if it is a verb or a nominalization).
\paragraph{Examples:} \hspace*{\fill}
\begin{itemize}
\item stood with her back against the wall
\item a water jug on the table
\end{itemize} 

{\small


\begin{tabular}{|l|l|}
\hline
{\bf Prep.} & {\bf Senses} \\ \hline
against&10(4) \\
on&1(1) 4(1c) 5(1d) \\
\hline\end{tabular}}
\paragraph{Comments:}   This may often overlap with \relation{Location}. It is the preferred label if there is the notion of one object being in physical contact with and supported by an surface or another object. Examples include {\em knelt on the cold stone floor} and {\em placed the jug on the table}.
\subsection{Possessor}
\paragraph{Definition:}  The governor of the preposition is something belonging to the object or an inherent quality of the object . This relation includes familial relations.
\paragraph{Examples:} \hspace*{\fill}
\begin{itemize}
\item a look about her
\item a black filly by Guldfuerst
\item his son by his third wife
\item son of a friend
\item a photograph of a bride
\item a few pounds on her.
\end{itemize} 

{\small


\begin{tabular}{|l|l|}
\hline
{\bf Prep.} & {\bf Senses} \\ \hline
about&5(3a)$^*$ \\
by&10(2e)$^*$ 9(2d)$^*$ \\
of&6(3) \\
on&6(1e)$^*$ \\
\hline\end{tabular}}
\paragraph{Comments:}   \begin{enumerate} \item Note that the \relation{Possessor} relation is largely represented in the training set via the preposition {\em of} in the sense of:6(3). \item This relation is different from the \relation{Attribute} relation, where the governor specifies the attribute or quality of the object. \end{enumerate}
\subsection{ProfessionalAspect}
\paragraph{Definition:}  This relation signifies a professional relationship between the governor (or the subject of the governor, if the governor is a verb) and the object of preposition, which is an employer, a profession, an institution or a business establishment.
\paragraph{Examples:} \hspace*{\fill}
\begin{itemize}
\item began performing at the university
\item tutor for the University
\item works in publishing
\item serve on committees
\item she is with Inland Revenue
\item bank with TSB
\end{itemize} 

{\small


\begin{tabular}{|l|l|}
\hline
{\bf Prep.} & {\bf Senses} \\ \hline
at&4(2b)$^*$ \\
for&4(3a) \\
in&8(6) \\
on&10(4)$^*$ \\
with&13(8) 14(8a)$^*$ \\
\hline\end{tabular}}
\paragraph{Comments:}   This is an infrequent relation that could have been merged into the \relation{Other} category, but has been separated because it forms a distinct cluster.
\subsection{Purpose}
\paragraph{Definition:}  The object of the preposition specifies the purpose (i.e., a result that is desired, intention or reason for existence) of the governor.
\paragraph{Examples:} \hspace*{\fill}
\begin{itemize}
\item networks for the exchange of information
\item tools for making the picture frame.
\end{itemize} 

{\small


\begin{tabular}{|l|l|}
\hline
{\bf Prep.} & {\bf Senses} \\ \hline
for&5(4) \\
\hline\end{tabular}}
\subsection{Recipient}
\paragraph{Definition:}  The object of the preposition identifies the person or thing receiving something.
\paragraph{Examples:} \hspace*{\fill}
\begin{itemize}
\item unkind to her
\item donated to the hospital
\end{itemize} 

{\small


\begin{tabular}{|l|l|}
\hline
{\bf Prep.} & {\bf Senses} \\ \hline
to&8(3) \\
\hline\end{tabular}}
\paragraph{Comments:}   The difference between a \relation{Recipient} and the \relation{Beneficiary} is that in the latter case, the recipient gets some advantage from the action or event.
\subsection{Separation}
\paragraph{Definition:}  The relation indicates separation or removal. The object of the preposition is the entity that is removed.
\paragraph{Examples:} \hspace*{\fill}
\begin{itemize}
\item the party was ousted from power
\item tear the door off its hinges
\item burden off my shoulders
\item I stay off alcohol
\item part with possessions
\end{itemize} 

{\small


\begin{tabular}{|l|l|}
\hline
{\bf Prep.} & {\bf Senses} \\ \hline
from&10(7) \\
off&4(3) 5(3a)$^*$ 6(3b)$^*$ 7(4)$^*$ \\
with&16(10) \\
\hline\end{tabular}}
\subsection{Source}
\paragraph{Definition:}  The object of the preposition indicates the provenance of the governor or the subject of the governor. This includes cases where the object is the place of origin, the source of information or the creator of an artifact.
\paragraph{Examples:} \hspace*{\fill}
\begin{itemize}
\item I am from Hackeney
\item paintings of Rembrandt
\item book by Hemmingway
\item information from books
\end{itemize} 

{\small


\begin{tabular}{|l|l|}
\hline
{\bf Prep.} & {\bf Senses} \\ \hline
by&4(1c) \\
from&1(1) 13(10) 2(1a) 4(3) \\
of&7(3a) \\
\hline\end{tabular}}
\paragraph{Comments:}   \begin{enumerate} \item While it might be possible to split this into multiple sub-relations (location, source of information and creator-creation), all but the first will end up being infrequent. \item Some figurative uses are included into this category, such as {\em turning them away from appeasement} and {\em stagger from crisis to crisis}. \end{enumerate}
\subsection{Species}
\paragraph{Definition:}  This expresses the relationship between a general category or type and the thing being specified which belongs to the category. The governor is a noun indicating the general category and the object is an instance of that category.
\paragraph{Examples:} \hspace*{\fill}
\begin{itemize}
\item the city of Prague
\item this type of book
\end{itemize} 

{\small


\begin{tabular}{|l|l|}
\hline
{\bf Prep.} & {\bf Senses} \\ \hline
of&10(5a) 9(5) \\
\hline\end{tabular}}
\subsection{StartState}
\paragraph{Definition:}  The object of the preposition indicates the state or condition that an entity has left.
\paragraph{Examples:} \hspace*{\fill}
\begin{itemize}
\item recovered from the disease
\item a growth from \$2.2 billion to \$2.4 billion
\end{itemize} 

{\small


\begin{tabular}{|l|l|}
\hline
{\bf Prep.} & {\bf Senses} \\ \hline
from&10(7)-1 6(4) 7(4a)$^*$ \\
\hline\end{tabular}}
\paragraph{Comments:}   \begin{enumerate} \item This relation requires some entity to change state, with the object of the preposition indicating the initial state before the transformation. \item This relation also includes cases where the preposition indicates the first element in a numeric or conceptual range. \end{enumerate} Some examples that are not \relation{StartState} are: \begin{inparaenum} \item {\em the ball fell from his hands}, which is a \relation{Source} relation because the object indicates the original physical {\em location} of the ball, \item {\em came back from a holiday}, which is also a \relation{Source} relation, and \item {\em a steady increase from June}, which is a \relation{Temporal} relation because the object of the preposition is a temporal expression. \end{inparaenum}
\subsection{Temporal}
\paragraph{Definition:}  The {\em object} of the preposition specifies the time of when an event occurs, either as an explicit temporal expression or by indicating another event as a reference. In some prepositional phrases, the governor may indicate a temporal expression. These do not express a \relation{Temporal} relation.
\paragraph{Examples:} \hspace*{\fill}
\begin{itemize}
\item shortly after Christmas
\item go to bed at nine o'clock
\item cooler at night
\item rest before dinner
\item open during the party
\item jailed for 12 years
\item the show will run from ten to two
\item met in 1985
\item reported on September 26
\end{itemize} 

{\small


\begin{tabular}{|l|l|}
\hline
{\bf Prep.} & {\bf Senses} \\ \hline
across&1(1)-1$^*$ \\
after&1(1) 2(1a)$^*$ 3(1b)$^*$ \\
at&2(2) 3(2a)$^*$ \\
before&1(1) \\
behind&7(4)$^*$ \\
between&2(2)$^*$ \\
by&14(3b)$^*$ 17(4)$^*$ 20(6)$^*$ \\
down&5(2)$^*$ \\
during&1(1) 2(1a) \\
for&12(10) \\
from&3(2) 5(3a) \\
in&3(2) 4(3)$^*$ \\
inside&5(2)$^*$ \\
into&1(1)-1$^*$ \\
of&18(9)$^*$ \\
on&17(8) 18(8a)$^*$ \\
over&14(5)$^*$ \\
through&11(4)$^*$ 7(2)$^*$ 9(2b) \\
to&4(1c)$^*$ \\
towards&3(1b)$^*$ \\
\hline\end{tabular}}
\subsection{Topic}
\paragraph{Definition:}  The object of the preposition denotes the subject or topic under consideration. In many cases, the preposition can be replaced by the phrase {\em on the subject of}. The governor of the preposition can be a verb or nominalization that implies an action or analysis or a noun that indicates an information store.
\paragraph{Examples:} \hspace*{\fill}
\begin{itemize}
\item thinking about you
\item book on careers
\item insight into what was involved
\item debate over unemployment
\end{itemize} 

{\small


\begin{tabular}{|l|l|}
\hline
{\bf Prep.} & {\bf Senses} \\ \hline
about&1(1) 2(1a)$^*$ \\
around&2(1a) \\
into&5(5)$^*$ \\
on&8(3) \\
over&16(7) \\
round&2(1a)$^*$ \\
\hline\end{tabular}}
\subsection{Via}
\paragraph{Definition:}  This is an infrequent relation where the object of the preposition indicates a mode of transportation or a path for travel. The governor can be action indicating movement or travel or a noun denoting passengers.
\paragraph{Examples:} \hspace*{\fill}
\begin{itemize}
\item traveling by bus
\item he is on his way
\item sleep on the plane
\item got on the train
\item go through the tube
\end{itemize} 

{\small


\begin{tabular}{|l|l|}
\hline
{\bf Prep.} & {\bf Senses} \\ \hline
by&8(2c) \\
on&14(7) 15(7a) 16(7b)$^*$ \\
onto&2(2)$^*$ \\
through&5(1d)-1 \\
\hline\end{tabular}}
\paragraph{Comments:}   This relation applies only when there is evidence of a physical travel or a mode of conveyance.

\section{Statistics}
\label{sec:statistics}
Table 1 shows the number of training and test examples for each
relation, obtained by applying the mapping defined in Section
\ref{sec:preposition-relations-list} to the training and test data of
the SemEval-2007 shared task.

Note that even though the label \relation{Other} is associated with
the highest number of senses, it has a comparatively small number of
examples associated with it. The relations that have fewer examples
than \relation{Other} have considerably fewer senses that define it,
indicating that they are more coherent classes.

\begin{table}[h!]
  \centering
  {\footnotesize
  \begin{tabular}{|l|c|c|}
    \hline
    {\bf Relation }                    & {\bf Train } & {\bf Test } \\
    \hline
    \relation{Activity}                & 63           & 39          \\
    \relation{Agent}                   & 367          & 159         \\
    \relation{Attribute}               & 510          & 266         \\
    \relation{Beneficiary}             & 205          & 105         \\
    \relation{Cause}                   & 591          & 289         \\
    \relation{Co-Particiants}          & 112          & 58          \\
    \relation{Destination}             & 1054         & 526         \\
    \relation{Direction}               & 909          & 441         \\
    \relation{EndState}                & 188          & 104         \\
    \relation{Experiencer}             & 116          & 54          \\
    \relation{Instrument}              & 565          & 290         \\
    \relation{Location}                & 3096         & 1531        \\
    \relation{Manner}                  & 457          & 245         \\
    \relation{MediumOfCommunication}   & 57           & 30          \\
    \relation{Numeric}                 & 113          & 48          \\
    \relation{ObjectOfVerb}            & 1801         & 882         \\
    \relation{Opponent/Contrast}       & 233          & 131         \\
    \relation{Other}                   & 72           & 42          \\
    \relation{PartWhole}               & 958          & 471         \\
    \relation{Participant/Accompanier} & 292          & 142         \\
    \relation{PhysicalSupport}         & 399          & 202         \\
    \relation{Possessor}               & 508          & 269         \\
    \relation{ProfessionalAspect}      & 45           & 22          \\
    \relation{Purpose}                 & 261          & 113         \\
    \relation{Recipient}               & 378          & 190         \\
    \relation{Separation}              & 345          & 172         \\
    \relation{Source}                  & 740          & 357         \\
    \relation{Species}                 & 394          & 198         \\
    \relation{StartState}              & 69           & 37          \\
    \relation{Temporal}                & 331          & 157         \\
    \relation{Topic}                   & 886          & 462         \\
    \relation{Via}                     & 61           & 26          \\
    \hline
    Total                              & 16176        & 8058        \\
    \hline

  \end{tabular}

  \caption{Number of training and test examples}}
  \label{tab:num-train-test}
\end{table}

\bibliographystyle{acl2012}
\bibliography{references}

\end{document}